\documentclass{article}

\usepackage{PRIMEarxiv}

\usepackage[utf8]{inputenc} % allow utf-8 input
\usepackage[T1]{fontenc}    % use 8-bit T1 fonts
\usepackage{hyperref}       % hyperlinks
\usepackage{url}            % simple URL typesetting
\usepackage{booktabs}       % professional-quality tables
\usepackage{amsfonts}       % blackboard math symbols
\usepackage{nicefrac}       % compact symbols for 1/2, etc.
\usepackage{microtype}      % microtypography
\usepackage{lipsum}
\usepackage{fancyhdr}       % header
\usepackage{graphicx}       % graphics
\usepackage{amsmath}
\graphicspath{{media/}}     % organize your images and other figures under media/ folder

\usepackage{pgfplots}
\pgfplotsset{compat=1.7}

\title{Forged Image Detection using SOTA Image Classification Deep Learning Methods for Image Forensics with Error Level Analysis}

\author{
    Raunak Joshi \\
      Department of IT \\
      University of Mumbai \\
      Mumbai, 400032, India \\
      \texttt{raunakjoshi.m@gmail.com} \\
    \And
  Abhishek Gupta \\
  Department of EXTC \\
  University of Mumbai \\
  Mumbai, 400032, India \\
  \texttt{abhishekgupta20001@gmail.com} \\
   \And
  Nandan Kanvinde \\
  Department of MCA \\
  TIMSCDR \\
  Mumbai, 400101, India \\
  \texttt{kanvindenandan81@gmail.com} \\
  \And
  \AND
  Pandharinath Ghonge \\
  Department of EXTC \\
  St. John College of Engineering and Management \\
  Palghar - 401404, India \\
  \texttt{pandharinathg@sjcem.edu.in} \\
}

\begin{document}
\maketitle

\begin{abstract}
The advancement in the area of computer vision has been brought using deep learning mechanisms. Image Forensics is one of the major areas of computer vision application. Forgery of images is sub-category of image forensics and can be detected using Error Level Analysis. Using such images as an input, this can turn out to be a binary classification problem which can be leveraged using variations of convolutional neural networks. In this paper we perform transfer learning with state-of-the-art image classification models over error level analysis induced CASIA ITDE v.2 dataset. The algorithms used are VGG-19, Inception-V3, ResNet-152-V2, XceptionNet and EfficientNet-V2L with their respective methodologies and results.
\end{abstract}

% keywords can be removed

\section{Introduction}
The area of computer vision \cite{Wiley2018ComputerVA} has excelled in terms of innovation and performance delivered by leveraging Deep Learning \cite{lecun2015deep}. The various tasks of computer vision are classification, object detection \cite{zhao2019object}, object counting \cite{guan2020understanding}, image segmentation \cite{minaee2021image,joshi2022refactoring} and many more. Classification \cite{Cormack1971ARO} can be termed as one of the most primordial tasks in computer vision. The task of classification is identification of an entity or object by prognosticating its appropriate label is done effectively using Convolutional Neural Networks \cite{lecun1995convolutional} abbreviated as CNN. The standard CNN gone under massive improvements in the latter period when ImageNet \cite{5206848} Large Scale Visual Recognition Challenge (ILSVRC) \cite{russakovsky2015imagenet} came into existence yielding many models that are currently state-of-the-art deep learning models for image classification. These models were later refined to a greater extent and turned out to be a base for many new models effectively. One sub-category in the area of classification is image forgery \cite{8203904}, where the deep learning can be used with binary classification \cite{kumari2017machine,kanvinde2022binary,gupta2022prediction} problem. The task will be simple as classifying whether the image is original or fake, hence leading to binary classification problem.

The image forgery can be termed as a sub-category of image forensics \cite{Piva2013AnOO} where there are many methods where Error Level Analysis \cite{7412439} is one of the commonly used techniques. The image when tampered performs compression which is used as the main point for Error Level Analysis abbreviated as ELA. The compression is obtained at 95\% of the input image and saved, which is later evaluated with the original image for differences. Images have grid squares which are thoroughly evaluated by ELA which yield errors due to successive saving of the image after editing. The forged section of the image is bright after ELA operation whereas non-tampered area is darker than all the parts of the image. The section highlights with lighter shades where tampering is done specifically. This data can be labelled and given to CNN for classification where the ELA filters over an image are input. Standard CNN application \cite{gupta2022detection} is very obvious and lack the necessary performance where the transfer learning \cite{zhuang2020comprehensive} can be performed with state-of-the-art models. The varied popular models can be used, viz. VGG \cite{simonyan2014very}, ResNet \cite{he2016deep,gupta2022residual}, InceptionNet \cite{szegedy2015going}, XceptionNet \cite{chollet2017xception} and EfficientNet \cite{tan2019efficientnet}. The appropriate data for such an application is CASIA dataset, where the CASIA ground-truth \cite{pham2019hybrid} is the first dataset that has distinct 8 classes of images. The tampering over this dataset was performed which yielded CASIA ITDE v.2 \cite{6625374} dataset having 2 distinct classes, authentic images and forged images. This paper is compendious comparison of state-of-the-art models using transfer learning over the CASIA ITDE v.2, which will be intricately explained in further sections of this paper.

\section{Methodology}
\subsection{VGG-19}
The VGG \cite{simonyan2014very} is one of the primarily used deep learning computer vision classification based algorithm which has 2 variants, 16 layered known as VGG-16 and 19 layered known as VGG-19. The network consists of blocks, where each block has 2-dimensional convolutional layers and 2-dimensional max-pooling \cite{gholamalinezhad2020pooling} layers. The VGG-19 has total 5 blocks, where first 2 blocks have 2 convolutional layers followed by one max-pooling layer. The remaining 3 blocks have 4 convolutional layers followed by max-pooling layer. Since the model was originally trained on imagenet \cite{5206848} data which has 1000 classes, we are using the same trained weights of the imagenet for transfer learning. We ignore the output layer of model and append 2-dimensional global average pooling \cite{lin2013network} layer followed by 2 layers. The first layer has 1024 hidden neurons with ReLU \cite{agarap2018deep} activation function, followed by the output layer that has 2 hidden neurons with softmax \cite{10.5555/2969830.2969856} activation function. Since we are dealing with binary classification problem, we could use 1 hidden neuron with sigmoid \cite{10.1016/S0893-6080(05)80129-7} activation function, but we wanted the output probabilities for inference examination in a through manner. The network for backward propagation \cite{rumelhart1985learning} uses binary cross-entropy loss function \cite{ruby2020binary} and Adam \cite{kingma2014adam} optimizer for loss optimization. The network has learned 20,551,746 total parameters which are approximately, 20.55 million parameters. If we remove the personally initialized layers the VGG-19 itself consists of roughly 20.02 million parameters.

\subsection{Inception-V3}
The computationally efficient network was definitely the InceptionNet \cite{szegedy2015going} consisting of 22 layers. The inception module was the main point of focus for the network that later made it a state-of-the-art model. The inception modules force the network to train many different filters in parallel fashion with concatenation for depth-wise aspect. This network over the period of time had many improvements, where batch normalization layers were used which is known as Inception-V2 \cite{ioffe2015batch} network. Later the network was made deeper with 42 layers which still turned out to be computationally efficient as compared to VGG. This modified network was Inception-V3 \cite{szegedy2016rethinking} that we have used, where we stripped of the output layers and applied 2-dimensional global average pooling \cite{lin2013network} followed by one 1024 hidden neuron ReLU \cite{agarap2018deep} activation layer and one 2 hidden neuron softmax \cite{10.5555/2969830.2969856} activation output layer. The loss function is binary cross-entropy and loss optimizer is Adam. The network has total 23,903,010 parameters, which approximately are around 23.90 million. The trainable parameters are 23,868,578 and non-trainable parameters are 34,432. 21,802,784 are the sole parameters of the Inception-V3 which are roughly 21.80 million respectively.

\subsection{ResNet}
The ResNet \cite{he2016deep} is termed as one of the most prominently used network for image classification tasks. The main problem that ResNet emphasized on was the degradation problem with convergence of the network. This is basically saturation of the accuracy with increasing depth of the network. For this the residual block\cite{gupta2022residual} was developed which uses a concept known as skip connection which abruptly skips the layers in between while back-propagation \cite{rumelhart1985learning} to avoid the saturation in loss optimization. The network we have used is ResNet with 152 layers. Now it has 2 versions, where version 1 uses non-linearity in the last layer and version 2 has removed it. We have used the ResNet 152 v2 \cite{he2016identity}, where we have removed the output layer and added 2-dimensional global average pooling \cite{lin2013network} with 1024 hidden neuron ReLU \cite{agarap2018deep} activation layer followed by 2 hidden neuron softmax \cite{10.5555/2969830.2969856} activation output layer. The loss function and optimizer is same as the earlier networks and it learns 60,431,874 parameters which are approximately 60.4 million parameters. After removing the custom added layers, the ResNet alone has 58,331,648 parameters.

\subsection{XceptionNet}
The XceptionNet \cite{chollet2017xception} is 71 layers deep extension network of Inception-V3 \cite{szegedy2016rethinking} that replaces the inception modules with depth-wise separable convolutions. The depth of the input is not only covered in this network but also spatial dimensions which makes it special which also covers the kernels that cannot be separated into further sections. We remove the output layer and add custom layers where first 2-dimensional global average pooling is applied. Followed by it is one layer with 1024 hidden neurons and ReLU activation function. Finally one output layer with 2 hidden neurons and softmax activation function is used. The loss function used is binary cross-entropy and loss optimizer used is Adam. The network learns total of 22,961,706 parameters which are approximately 22.96 million. The non-trainable parameters are 54,528 only and 22,907,178 are trainable. The parameters excluding the custom defined layers are 20,861,480, which roughly are around 20.86 million respectively.

\subsection{EfficientNet-V2L}
The need of efficiency in training and computation of the networks has risen in a broader perspective for which EfficientNet \cite{tan2019efficientnet} drives as a good example. The scaling of the base features of the input like width, height and depth is done arbitrarily which has been made uniform with EfficientNet. The compound scaling is the main feature and many variation of it were made where advancement also reached the version 2. We have used EfficientNet-V2L \cite{tan2021efficientnetv2} which is large network with over 110 million parameters. The network we have trained has 119,060,642 parameters which are around 119 million roughly where the non-trainable parameters are around 500k and rest are trainable. The network without our layers is 117,746,848 parameters which is approximately 117.74 million parameters. The last layer of the EfficientNet-V2L is replaced and 2-dimensional global average pooling is applied. Followed by it is one dense layer with 1024 hidden neurons and ReLU activation function. The final output layer is applied with 2 hidden neurons and softmax activation function. The loss function used is binary cross-entropy and adam loss optimizer for loss optimization. 

\section{Results}
\subsection{Training and Validation Accuracy}

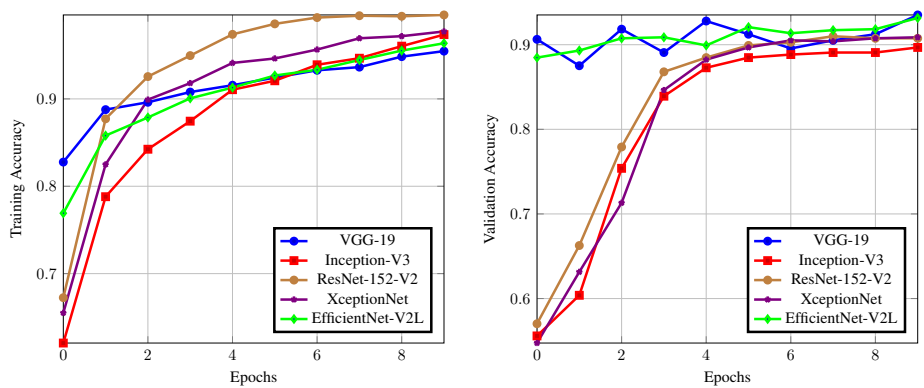
\begin{figure*}[htbp]
    \centering
    \begin{tabular}{c c}
        \begin{tikzpicture}[scale=0.6]
            \begin{axis}[xlabel={Epochs},ylabel ={Training Accuracy},enlargelimits=false,
            grid=both,
            scale only axis=true,legend pos=south east,style={ultra thick}, axis line style={ultra thin}]
            \addplot+[blue] table[x=Epochs,y=vgg,col sep=comma]{plots/accuracy.csv}; 
            \addplot+[red] table[x=Epochs,y=inception,col sep=comma]{plots/accuracy.csv}; 
            \addplot+[brown] table[x=Epochs,y=res,col sep=comma]{plots/accuracy.csv};
            \addplot+[violet] table[x=Epochs,y=xception,col sep=comma]{plots/accuracy.csv};
            \addplot+[green] table[x=Epochs,y=efficient,col sep=comma]{plots/accuracy.csv};
            \addlegendentry{VGG-19}
            \addlegendentry{Inception-V3}
            \addlegendentry{ResNet-152-V2}
            \addlegendentry{XceptionNet}
            \addlegendentry{EfficientNet-V2L}
            \end{axis}
        \end{tikzpicture}
        &
        \begin{tikzpicture}[scale=0.6]
            \begin{axis}[xlabel = {Epochs},ylabel = {Validation Accuracy},enlargelimits=false,
            grid=both,
            scale only axis=true,legend pos=south east,style={ultra thick}, axis line style={ultra thin}]
            \addplot+[blue] table[x=Epochs,y=vgg,col sep=comma]{plots/val_accuracy.csv}; 
            \addplot+[red] table[x=Epochs,y=inception,col sep=comma]{plots/val_accuracy.csv}; 
            \addplot+[brown] table[x=Epochs,y=res,col sep=comma]{plots/val_accuracy.csv};
            \addplot+[violet] table[x=Epochs,y=xception,col sep=comma]{plots/val_accuracy.csv};
            \addplot+[green] table[x=Epochs,y=efficient,col sep=comma]{plots/val_accuracy.csv};
            \addlegendentry{VGG-19}
            \addlegendentry{Inception-V3}
            \addlegendentry{ResNet-152-V2}
            \addlegendentry{XceptionNet}
            \addlegendentry{EfficientNet-V2L}
            \end{axis}
        \end{tikzpicture}
    \end{tabular}
    \caption{Training Accuracy and Validation Accuracy}
    \label{fig:a}
\end{figure*}

The below figure \ref{fig:a} gives a general overview of the training and validation accuracy for all the models used in the implementation. The highest accuracy in the last epoch of training accuracy for VGG-19 is 95.46\% and validation accuracy is 93.51\% which is very effectively generalized. Similarly the training accuracy for Inception-V3 is 97.35\% whereas the validation accuracy is 89.67\% which makes model less generalized but a better training accuracy than VGG-19. The training accuracy for ResNet152-V2 is 99.6\% and validation accuracy is 90.75\%. The training accuracy for XceptionNet is 97.71\% and validation accuracy is 90.87\%. Finally the training accuracy for EfficientNet-V2L is 96.36\% and validation accuracy is 93.15\% making it generalized and effective as compared to other networks yet this is not sufficient metrics for stating an inference.

\subsection{Training and Validation Loss}

\begin{figure*}[htbp]
    \centering
    \begin{tabular}{c c}
        \begin{tikzpicture}[scale=0.6]
            \begin{axis}[xlabel={Epochs},ylabel ={Training Loss},enlargelimits=false,
            grid=both,
            scale only axis=true,legend pos=north east,style={ultra thick}, axis line style={ultra thin}]
            \addplot+[blue] table[x=Epochs,y=vgg,col sep=comma]{plots/loss.csv}; 
            \addplot+[red] table[x=Epochs,y=inception,col sep=comma]{plots/loss.csv}; 
            \addplot+[brown] table[x=Epochs,y=res,col sep=comma]{plots/loss.csv};
            \addplot+[violet] table[x=Epochs,y=xception,col sep=comma]{plots/loss.csv};
            \addplot+[green] table[x=Epochs,y=efficient,col sep=comma]{plots/loss.csv};
            \addlegendentry{VGG-19}
            \addlegendentry{Inception-V3}
            \addlegendentry{ResNet-152-V2}
            \addlegendentry{XceptionNet}
            \addlegendentry{EfficientNet-V2L}
            \end{axis}
        \end{tikzpicture}
        &
        \begin{tikzpicture}[scale=0.6]
            \begin{axis}[xlabel = {Epochs},ylabel = {Validation Loss},enlargelimits=false,
            grid=both,
            scale only axis=true,legend pos=north east,style={ultra thick}, axis line style={ultra thin}]
            \addplot+[blue] table[x=Epochs,y=vgg,col sep=comma]{plots/val_loss.csv}; 
            \addplot+[red] table[x=Epochs,y=inception,col sep=comma]{plots/val_loss.csv}; 
            \addplot+[brown] table[x=Epochs,y=res,col sep=comma]{plots/val_loss.csv};
            \addplot+[violet] table[x=Epochs,y=xception,col sep=comma]{plots/val_loss.csv};
            \addplot+[green] table[x=Epochs,y=efficient,col sep=comma]{plots/val_loss.csv};
            \addlegendentry{VGG-19}
            \addlegendentry{Inception-V3}
            \addlegendentry{ResNet-152-V2}
            \addlegendentry{XceptionNet}
            \addlegendentry{EfficientNet-V2L}
            \end{axis}
        \end{tikzpicture}
    \end{tabular}
    \caption{Training Loss and Validation Loss}
    \label{fig:b}
\end{figure*}
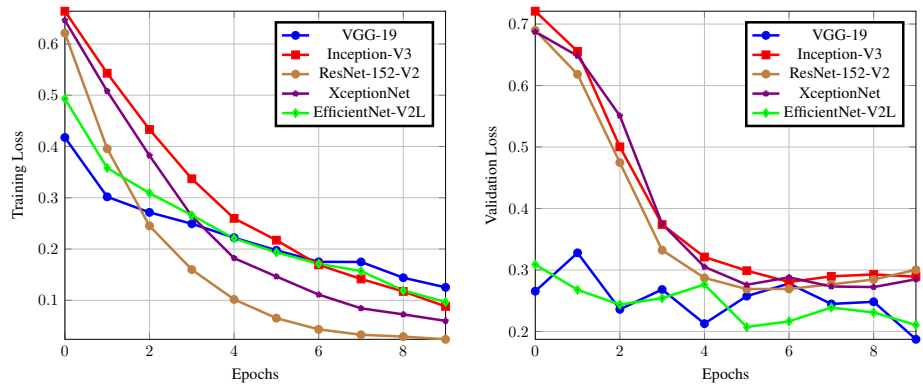

The figure \ref{fig:b} gives a graphical depiction of the models converging throughout the epochs. The training loss for VGG-19 is 12.25\% in the last epochs whereas the validation loss is 18.72\%. The training loss for Inception-V3 is 8.8\% and validation loss is 28.91\%. The training loss for ResNet-152V2 is 2.39\% while the validation loss is 30.03\% effectively. The training loss for XceptionNet is 5.98\% whereas the validation loss is 28.48\%. Finally the training loss for EfficientNet-V2L is 9.67\% and validation loss is 21.05\% respectively. The generalization can be seen better with VGG-19 but highest training loss convergence can be seen with ResNet-152V2. EfficientNet-V2L attains a middle ground effectively.

\subsection{Precision}

The precision \cite{powers2020evaluation} in simple terms is positive sample for classification in model accuracy. The better explanation can be given as number of positive samples classified with respect to total number of samples. The table \ref{tab:a} below gives an explanation of precision scores for all the respective models. 

\begin{table}[htbp]
\caption{Precision Score}
\centering
\begin{tabular}{ll}
\toprule
\textbf{Model} & \textbf{Precision Score} \\
\midrule
\textbf{VGG-19} & \textbf{98.25\%} \\
Inception-V3 & 94.11\% \\
ResNet-152-V2 & 90.38\% \\
XceptionNet & 94.61\% \\
EfficientNet-V2L & 85.20\% \\
\bottomrule
\end{tabular}
\label{tab:a}
\end{table}

The precision obtained for VGG-19 is highest which is an indication of higher number of positive samples classified as per the total samples effectively. The precision alone though is not the perfect metric for deriving the final inference.

\subsection{Recall}

The recall \cite{powers2020evaluation} is number of positive samples classified appropriately with respect to total number of positive samples. The table \ref{tab:b} gives the recall scores for all the algorithms used. 

\begin{table}[htbp]
\caption{Recall Score}
\centering
\begin{tabular}{ll}
\toprule
\textbf{Model} & \textbf{Recall Score} \\
\midrule
VGG-19 & 79.34\% \\
Inception-V3 & 82.62\% \\
ResNet-152-V2 & 88.26\% \\
XceptionNet & 86.61\% \\
\textbf{EfficientNet-V2L} & \textbf{94.60\%} \\
\bottomrule
\end{tabular}
\label{tab:b}
\end{table}

The Recall score is highest for EfficientNet-V2L which is was directly lowest in terms of precision. This is the reason recall also cannot be used for final inference judgement.

\subsection{RoC and AUC}

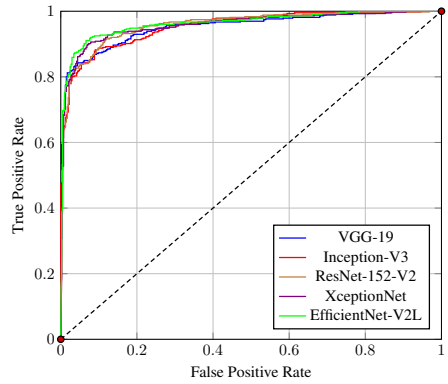
\begin{figure}[htbp]
    \centering
    \begin{tikzpicture}[scale=0.6]
        \begin{axis}[xlabel={False Positive Rate},ylabel={True Positive Rate},enlargelimits=false,
        grid=both,
        scale only axis=true,legend pos=south east,style={thick}, axis line style={ultra thin}]
        \addplot+[blue, no markers] table[x=fpr_vgg,y=tpr_vgg,col sep=comma]{plots/rocauc.csv}; 
        \addplot+[red, no markers] table[x=fpr_inc,y=tpr_inc,col sep=comma]{plots/rocauc.csv}; 
        \addplot+[brown, no markers] table[x=fpr_res,y=tpr_res,col sep=comma]{plots/rocauc.csv};
        \addplot+[violet, no markers] table[x=fpr_xcp,y=tpr_xcp,col sep=comma]{plots/rocauc.csv};
        \addplot+[green, no markers] table[x=fpr_eff,y=tpr_eff,col sep=comma]{plots/rocauc.csv};
        \addplot+[black] coordinates { (0,0) (1,1) };
        \addlegendentry{VGG-19}
        \addlegendentry{Inception-V3}
        \addlegendentry{ResNet-152-V2}
        \addlegendentry{XceptionNet}
        \addlegendentry{EfficientNet-V2L}
        \end{axis}
    \end{tikzpicture}
    \caption{RoC and AUC}
    \label{fig:c}
\end{figure}

The Receiver operating characteristic \cite{Fawcett2006AnIT} curve is specifically for binary classification that gives the performance of classification at all the thresholds ranging from 0 to 1. It uses false and true positive rates for evaluation of the RoC curve which the abbreviation basically. The area under curve is also its crucial aspect where the entire area covering the RoC and signifies the separability of classes. It is difficult to make a judgement from figure \ref{fig:c} since the predictions are merging, for which zooming into the depth can help derive some intuition. 

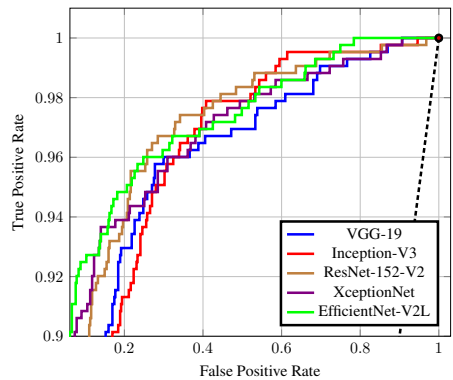
\begin{figure}[htbp]
    \centering
    \begin{tikzpicture}[scale=.6]
        \begin{axis}[xlabel={False Positive Rate},ylabel={True Positive Rate},enlargelimits=false,
        grid=both, ymax=1.01, ymin=0.9, xmax=1.03,
        scale only axis=true,legend pos=south east,style={ultra thick}, axis line style={ultra thin}]
        \addplot+[blue, no markers] table[x=fpr_vgg,y=tpr_vgg,col sep=comma]{plots/rocauc.csv}; 
        \addplot+[red, no markers] table[x=fpr_inc,y=tpr_inc,col sep=comma]{plots/rocauc.csv}; 
        \addplot+[brown, no markers] table[x=fpr_res,y=tpr_res,col sep=comma]{plots/rocauc.csv};
        \addplot+[violet, no markers] table[x=fpr_xcp,y=tpr_xcp,col sep=comma]{plots/rocauc.csv};
        \addplot+[green, no markers] table[x=fpr_eff,y=tpr_eff,col sep=comma]{plots/rocauc.csv};
        \addplot+[black] coordinates { (0,0) (1,1) };
        \addlegendentry{VGG-19}
        \addlegendentry{Inception-V3}
        \addlegendentry{ResNet-152-V2}
        \addlegendentry{XceptionNet}
        \addlegendentry{EfficientNet-V2L}
        \end{axis}
    \end{tikzpicture}
    \caption{Zoomed RoC with AUC}
    \label{fig:d}
\end{figure}

\begin{figure}[htbp]
    \centering
    \begin{tikzpicture}[scale=0.6]
        \begin{axis}[xlabel={False Positive Rate},ylabel={True Positive Rate},enlargelimits=false,
        grid=both, ymax=0.97, ymin=0.94,
        scale only axis=true,legend pos=south east,style={ultra thick}, axis line style={ultra thin}]
        \addplot+[blue, no markers] table[x=fpr_vgg,y=tpr_vgg,col sep=comma]{plots/rocauc.csv}; 
        \addplot+[red, no markers] table[x=fpr_inc,y=tpr_inc,col sep=comma]{plots/rocauc.csv}; 
        \addplot+[brown, no markers] table[x=fpr_res,y=tpr_res,col sep=comma]{plots/rocauc.csv};
        \addplot+[violet, no markers] table[x=fpr_xcp,y=tpr_xcp,col sep=comma]{plots/rocauc.csv};
        \addplot+[green, no markers] table[x=fpr_eff,y=tpr_eff,col sep=comma]{plots/rocauc.csv};
        \addplot+[black] coordinates { (0,0) (1,1) };
        \addlegendentry{VGG-19}
        \addlegendentry{Inception-V3}
        \addlegendentry{ResNet-152-V2}
        \addlegendentry{XceptionNet}
        \addlegendentry{EfficientNet-V2L}
        \end{axis}
    \end{tikzpicture}
    \caption{Super Zoomed RoC with AUC in range of 94\% to 97\% Y-axis}
    \label{fig:e}
\end{figure}
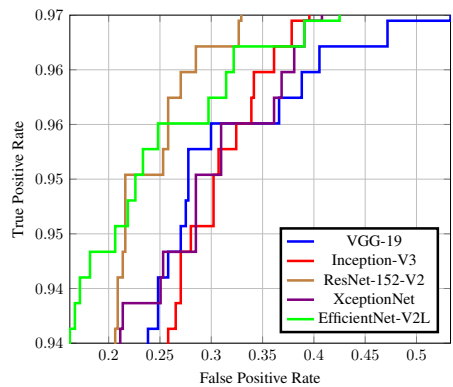

The figure \ref{fig:d} shows the pivots every algorithm made after a stagnant set of predictions. All the algorithms eventually reach the peak that is 1 value. The EfficientNet-V2L has it around 96\% as seen in figure \ref{fig:e} clearly. Big leap for Inception-V3 is at 95\% approximately. For ResNet-152-V2 it is somewhere above 95\% approximately. The VGG-19 turns just before 96\% and XceptionNet clearly at 96\%. These values are an indication of the learning procedures adopted for every classification threshold and area under their curve.

\subsection{F Measure}
All the metrics are very contradictory for making a judgement for the best algorithm. The precision indicates that VGG-19 is better whereas the EfficientNet-V2L is highest for recall. These both inferences are very far away from each other and proper metric is required. The averaging system for this is required where the F Measure also known as F-Score \cite{powers2020evaluation} is used. 

\begin{table}[htbp]
\caption{F-Measure for all the used Models}
\centering
\begin{tabular}{ll}
\toprule
\textbf{Model} & \textbf{F-Measure} \\
\midrule
VGG-19 & 87.79\% \\
Inception-V3 & 88\% \\
ResNet-152-V2 & 89.31\% \\
\textbf{XceptionNet} & \textbf{90.44\%} \\
EfficientNet-V2L & 89.65\% \\
\bottomrule
\end{tabular}
\label{tab:c}
\end{table}

The F measure has 1 as the $\beta$ value where the precision and recall are in the balanced state for metric measuring. If the $\beta$ value is 0.5 or 2, the metric inclines towards precision and recall respectively. These values can be represented with a table for comparison. The table \ref{tab:c} is the depiction of F measure for all the models used where XceptionNet outperforms all the models. Now this is most different observation and least expected. Since one relies on half measures, the results can turn out to be very varied but the least expected always turns out to be correct.

\section{Conclusion}
The area of image forgery is a potential application which in this paper has been leveraged using state-of-the-art deep learning image classification models with transfer learning. The dataset we used was CASIA ITDE v.2 which is a binary classification dataset that differentiates between authentic and tampered images. After performing the error level analysis from image forensics, the variety of famous models were used. Making an inference for the best model was a difficult task indeed due to results section variability yet the right judgement was attained. This paper does open potential thought for transfer learning models and wide variety of applications in the area of deep learning for computer vision. The limitations of the paper are only bounded and entitled to subtle differences between various state-of-the-art models that are very well explained throughout the result section of the paper. The distinction and clarity of reaching an inference in the paper is bounded to the methods of proving a point and we hope with our best belief that some more extensions with more better results can be obtained in future work.

%Bibliography
\bibliographystyle{unsrt}  
\bibliography{references}

\end{document}